\documentclass[sigconf]{acmart}
\AtBeginDocument{%
  }

\setcopyright{acmlicensed}
\copyrightyear{2018}
\acmYear{2018}
\acmDOI{XXXXXXX.XXXXXXX}
\acmConference[Conference acronym 'XX]{Make sure to enter the correct
  conference title from your rights confirmation email}{June 03--05,
  2018}{Woodstock, NY}
\acmISBN{978-1-4503-XXXX-X/2018/06}

\renewcommand\footnotetextcopyrightpermission[1]{}
\setcopyright{none}
\setcopyright{none}
\acmConference{}{}{}
\acmBooktitle{}
\acmPrice{}
\acmDOI{}
\acmISBN{}
\renewcommand\footnotetextcopyrightpermission[1]{}
\usepackage{booktabs}
\usepackage{multirow}
\usepackage{makecell}
\usepackage{graphicx}
\usepackage{array}
\usepackage{amsmath}
\usepackage{bm}
\usepackage{placeins}
\usepackage{dblfloatfix}
\usepackage[utf8]{inputenc}
\usepackage{textgreek}
\newcommand{\best}[1]{\bm{#1}}
\newcommand{\second}[1]{\underline{#1}}
\newcommand{\NA}{\textemdash}
\usepackage{booktabs}
\usepackage{float}
\begin{document}

\title{CLEAN: Psychometrically Consistent Incremental Cognitive Diagnosis under Concept-Space Expansion via Architectural Isolation}

\author{Tao He}
\affiliation{%
  \institution{Faculty of Education, Shenzhen University}
  \city{Shenzhen}
  \country{China}}
\email{tao.he2016@gmail.com}

\author{Jinxing Xiang}
\affiliation{%
  \institution{Faculty of Education, Shenzhen University}
  \city{Shenzhen}
  \country{China}}
\email{2510185003@mails.szu.edu.cn}

\author{Fan Jiang}
\correspondingauthor
\authornote{Corresponding author.}
\affiliation{%
  \institution{School of Education Science, Guangdong Polytechnic Normal University}
  \city{Guangzhou}
  \country{China}}
\email{fanjiang@zjnu.edu.cn}


\renewcommand{\shortauthors}{He et al.}

\begin{abstract}
Cognitive diagnosis (CD) is a fundamental task in intelligent education that profiles learner proficiency over knowledge concepts. In real-world learning platforms, newly added items continually introduce previously unseen concepts, necessitating dynamic expansion of the underlying concept space. Yet existing incremental CD models assume a fixed concept space, allowing gradients from new items to overwrite historical pathways and induce catastrophic forgetting. More critically, these methods rely solely on soft constraints to preserve historical diagnoses. Such constraints may fail to satisfy the requirement of diagnostic invariance after incremental updates, a requirement known as psychometric consistency in cognitive diagnosis. Therefore, we propose CLEAN (Continual Learning with Expandable and Architecturally Isolated Networks), a novel incremental CD framework supporting concept-space expansion while providing structural guarantees for pointwise invariance of historical diagnoses. Specifically, CLEAN first introduces a strict topological bipartition protocol, freezes historical diagnostic functions and applies deterministic orthogonal column masking to sever gradient interference. Second, to accommodate concept expansion, expandable full-rank branches with micro-variance initialization are deployed to learn novel concepts. Finally, to verify that this architectural design achieves invariance by construction, we formalize Representation Drift (RD) to quantify the perturbation of historical traits. Extensive experiments on three large-scale educational datasets demonstrate that CLEAN achieves zero RD, preserving old-item metrics identically to static anchors through architectural isolation while remaining competitive with or superior to strong continual-learning baselines on new items.

\end{abstract}




\maketitle
\pagestyle{plain}
\thispagestyle{plain}

\section{Introduction}
Cognitive diagnosis (CD) underpins personalized learning in intelligent tutoring systems (ITS), inferring each learner's proficiency over a structured space of knowledge concepts~\cite{wang_survey_cd_2024}. Classical models such as IRT~\cite{reckase_mirt_2009}, DINA~\cite{de_la_torre_dina_2009}, and NCDM~\cite{wang_ncdm_2020} are transductive: learners and items are represented by free parameters, so admitting a new entity requires global retraining. Generative diagnosis removes this bottleneck by replacing entity-specific embeddings with a shared Generative Diagnosis Function (GDF), which maps a sparse response vector to latent traits in a single forward pass~\cite{li_gncdm_2026}. Both paradigms, however, rest on the same implicit assumption that the concept space remains fixed.

\begin{figure*}[t]
  \centering
  \includegraphics[width=\textwidth]{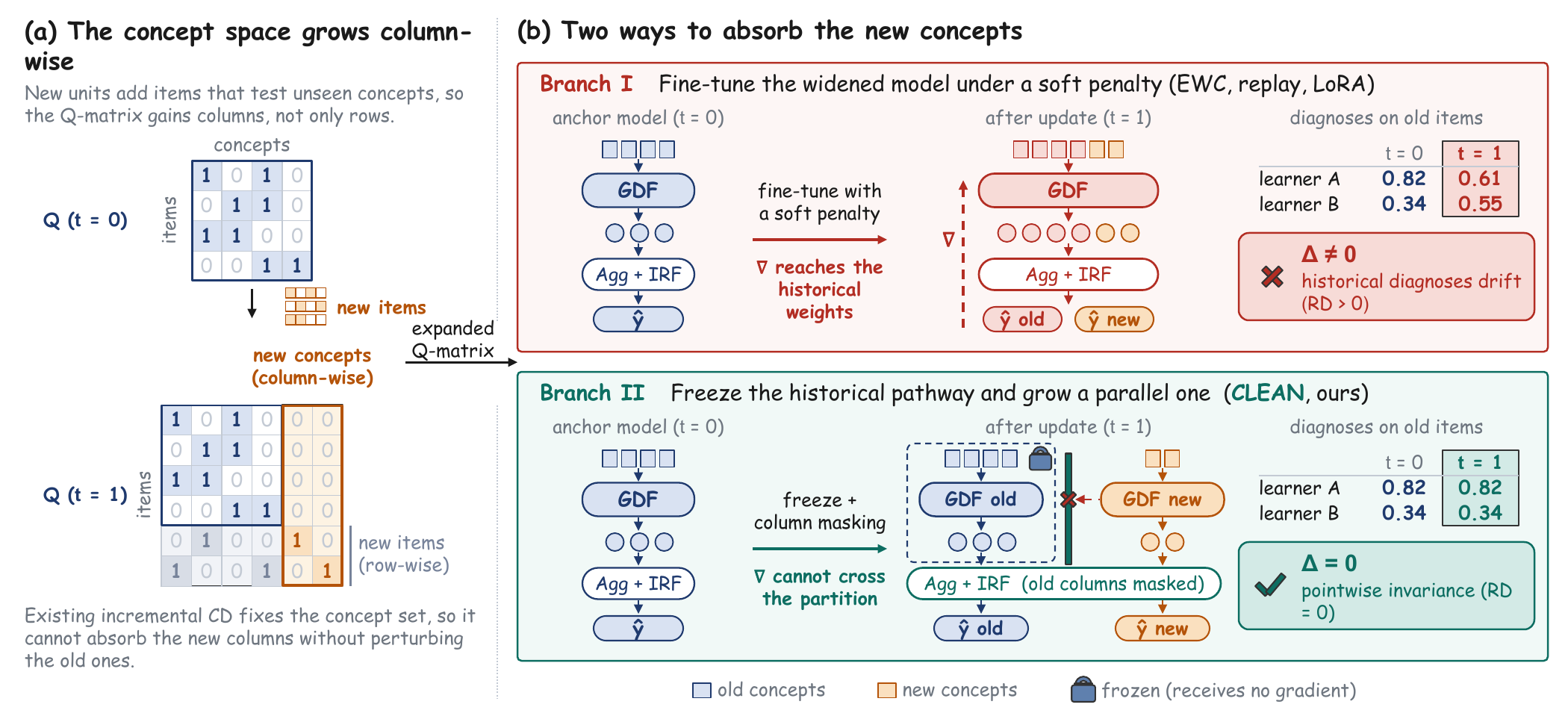}
  \caption{Incremental update of cognitive diagnosis models.
  (a) New units expand the $Q$-matrix column-wise, not merely row-wise.
  (b) Branch~I keeps every weight trainable, so gradients from new items
  reach the historical pathway and old-item diagnoses drift ($\mathrm{RD}>0$);
  Branch~II (CLEAN) freezes that pathway and masks the shared columns,
  making historical diagnoses invariant pointwise ($\mathrm{RD}=0$).}
  \label{fig:incremental-update}
\end{figure*}

Deployed platforms violate this assumption continually. As curricula extend and new instructional units are introduced, incoming items may involve previously unseen concepts and expand the $Q$-matrix column-wise rather than merely row-wise, so the diagnostic topology itself must expand. Existing incremental CD studies remain within the row-wise regime: ICD accommodates new learners and items but explicitly fixes the concept set~\cite{tong_icd_2022}, while other studies infer $Q$-vectors for new items~\cite{qin_qmatrix_incremental_2015}, diagnose newly registered learners~\cite{liu_icdm_2024}, or adapt to shifting task distributions~\cite{wu_metacd_2025}. Widening the trait layer and fine-tuning is the obvious fallback, yet it allows gradients from new items to overwrite the pathways responsible for generating historical diagnoses, causing catastrophic forgetting~\cite{kirkpatrick_ewc_2017} (Branch~I in Figure~\ref{fig:incremental-update}).

% \begin{figure}[t]
%   \centering
%   \includegraphics[width=\linewidth]{fig1}
%   \caption{The flowchart of incremental update for CDMs. Soft constraints provide only approximate stability and may still lead to baseline collapse, whereas architectural isolation preserves historical diagnoses through structural separation.}
%   \Description{The flowchart compares soft-constrained and architecture-isolated incremental updates for cognitive diagnosis models.}
%   \label{fig:incremental-update}
% \end{figure}

Continual learning offers several remedies, but they are calibrated to the wrong criterion. Regularization and replay~\cite{kirkpatrick_ewc_2017,li_lwf_2016,buzzega_derpp_2020} reduce drift rather than eliminate it, and replay further requires retaining historical responses, which privacy and storage constraints often preclude. Parameter-isolation methods such as PNN~\cite{rusu_pnn_2016} freeze prior columns yet reconnect them through dense lateral pathways, leaving historical and newly introduced diagnostic pathways entangled. More fundamentally, these methods certify retention by aggregate accuracy on old tasks, whereas a CD model is a measurement instrument: if a learner's reported mastery of fractions shifts merely because trigonometry is introduced into the platform, remediation is misdirected even when average accuracy remains unchanged. CD therefore requires a strict form of psychometric consistency related to measurement invariance~\cite{vandenberg_measurement_invariance_2000}, such that historical diagnoses must remain unchanged after incremental updates for the same response input. No soft penalty can guarantee this invariance.

Our key observation is that such invariance need not be enforced through optimization. The expert-annotated $Q$-matrix already specifies which concepts each item measures, allowing this topology to be compiled directly into the network. We therefore propose CLEAN (Continual Learning with Expandable, Architecturally Isolated Networks). Its Strict Topological Bipartition (STB) protocol reorders the $Q$-matrix into a block-triangular form in which old items never reference new concepts. Guided by it, CLEAN freezes historical GDFs, applies deterministic orthogonal column masking, and freezes the shared aggregation bias, so incremental gradients cannot reach historical pathways, while expandable full-rank or low-rank~\cite{hu_lora_2022} branches capture new concepts (Branch~II in Figure~\ref{fig:incremental-update}). Invariance thus becomes a property of the computational graph rather than an optimization objective.

The contributions of this work are summarized as follows:

\begin{itemize}
\item We formulate incremental CD under joint item- and concept-space expansion, and define Representation Drift (RD) to quantify perturbation of historical traits, taking $\mathrm{RD}=0$ as one important
operational criterion for psychometric consistency.

\item CLEAN compiles the $Q$-matrix into an isolation topology via STB and combines frozen historical pathways, orthogonal column masking, and micro-variance-initialized expandable branches, yielding invariance by construction, with a LoRA variant for constrained parameter budgets.

\item On Junyi, ASSISTments, and Math1, spanning dense to highly sparse responses, CLEAN attains $\mathrm{RD}=0$ and exactly retains anchor-level old-item metrics, while no baseline surpasses it on new items without either replaying stored responses or incurring substantial representation drift.
\end{itemize}

\section{RELATED WORK}
Concept-space expansion imposes two requirements that we use below to position
existing work: (i) \emph{column-wise expansion} of the Q-matrix when previously
unseen concepts arrive, and (ii) \emph{pointwise invariance} of historical
diagnoses after such expansion, as demanded by psychometric
consistency~\cite{vandenberg_measurement_invariance_2000}.

\noindent\textbf{Cognitive Diagnosis Models.}
By parameterization, CDMs fall into transductive and generative
families~\cite{wang_survey_cd_2024,li_gncdm_2026}. Transductive models assign
independent parameters to each learner and item, ranging from psychometric
formulations such as IRT~\cite{hambleton2013item}, MIRT~\cite{reckase_mirt_2009} and the Q-matrix-based
DINA~\cite{de_la_torre_dina_2009}, to neural and structure-aware extensions
including NCDM~\cite{wang_ncdm_2020}, DIRT~\cite{cheng_dirt_2019}, RCD~\cite{gao_rcd_2021}, HierCDF~\cite{li_hiercdf_2022}, and KSCD~\cite{ma_kscd_2022}.
Because these parameters are entity-specific, every newly arriving learner or
item triggers global retraining. Generative models remove this bottleneck by
mapping response logs to proficiency through shared diagnostic functions,
enabling immediate inference for unseen learners~\cite{li_gncdm_2026}. Both
families, however, are defined over a fixed concept space and therefore meet
neither requirement.

\noindent\textbf{Incremental Cognitive Diagnosis.}
Existing incremental work expands the diagnostic system along entities rather
than concepts. Qin et al.~\cite{qin_qmatrix_incremental_2015} estimate the
Q-vectors of new items from correctly specified base items, which appends rows
to the Q-matrix while leaving its columns fixed. ICD~\cite{tong_icd_2022}, the
work closest to ours, accommodates new learners and items through a deep trait
network and a stability-oriented update rule, yet explicitly assumes an
unchanged concept set. ICDM~\cite{liu_icdm_2024} targets sparse response logs of
newly registered learners, and MetaCD~\cite{wu_metacd_2025} combines
meta-learning with parameter protection under shifting data distributions. None
of these methods supports requirement (i), and each preserves historical
diagnoses only approximately, through soft stability terms rather than by
construction.

\noindent\textbf{Continual Learning and the Measurement of Forgetting.}
Continual learning mitigates catastrophic
forgetting~\cite{goodfellow_catastrophic_2014,parisi_continual_2019} along three
lines. Regularization penalizes changes to important parameters or distills
previous outputs~\cite{kirkpatrick_ewc_2017,li_lwf_2016}; replay rehearses
stored samples or
logits~\cite{rebuffi_icarl_2017,lopezpaz_gem_2017,chaudhry_agem_2019,buzzega_derpp_2020,boschini_xder_2023},
which presumes continued access to historical data; architecture-based methods
allocate task-specific parameters, either as full columns with lateral
connections, as in PNN~\cite{rusu_pnn_2016}, or as low-rank adapters constrained
by soft orthogonality, protected subspaces, and dual branches, as in the LoRA
family~\cite{hu_lora_2022,zhang_clora_2025,liang_inflora_2024,he_cllora_2025,xiong_oplora_2026}.
The first two lines suppress drift without eliminating it, whereas the third is
driven by task identity and is therefore blind to the concept-level topology
encoded in the Q-matrix. Standard evaluation masks this gap: forgetting is
typically reported through aggregate old-task statistics such as average
accuracy or forgetting scores~\cite{chaudhry_rwalk_2018,de_lange_survey_2022},
and drift-aware methods estimate representation shift in order to compensate for
it after the fact~\cite{yu_semantic_drift_2020}. A model may thus retain its
aggregate accuracy on historical items while silently altering individual
learners' proficiency on existing concepts, which is precisely what psychometric
consistency forbids. Our CLEAN closes both gaps by making historical invariance a
structural property of the architecture rather than an optimization objective.

\begin{figure*}[t]
  \centering
   \scalebox{1}[1.0]{%
  \includegraphics[width=0.9\textwidth]{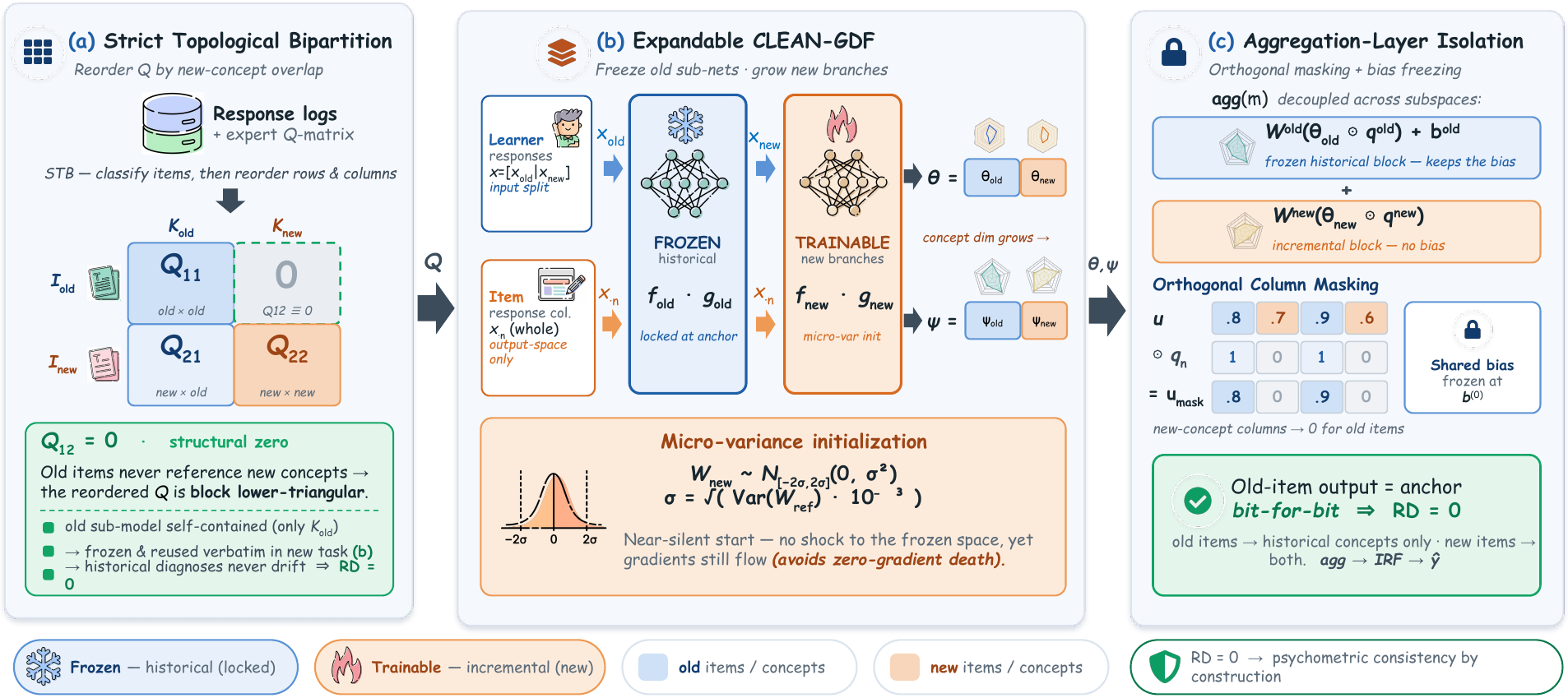}%
  }
  \caption{Overview of the CLEAN framework at the incremental training stage.}
  \Description{An overview of the CLEAN framework, including topological reordering, the generative diagnostic functions, and the aggregation layer.}
  \label{fig:clean-framework}
\end{figure*}

\section{PRELIMINARY}

\subsection{\textbf{Generative Cognitive Diagnosis Backbone}}
Let $\mathcal{U}=\{u_1,\ldots,u_N\}$,
$\mathcal{I}=\{I_1,\ldots,I_M\}$, and
$\mathcal{K}=\{k_1,\ldots,k_T\}$ denote the sets of learners,
items, and knowledge concepts, respectively. Their associations
are specified by an expert-defined $Q$-matrix
$Q\in\{0,1\}^{M\times T}$, where $Q_{mt}=1$ indicates that
item $I_m$ requires concept $k_t$~\cite{de_la_torre_dina_2009}.
Learner responses form a sparse matrix
$R\in\{0,1,-1\}^{N\times M}$, where $r_{im}\in\{0,1\}$
denotes an observed response and $-1$ denotes a missing interaction.

\textbf{Generative Diagnosis Function (GDF).}
Unlike transductive CDMs that represent learner and item profiles
as independent parameters, we adopt the generative diagnosis
paradigm~\cite{li_gncdm_2026}, which generates concept-aligned
traits from response logs:

\begin{equation}
\theta_i
=
\operatorname{GDF}(R_{i\cdot};\phi_{\theta}),
\qquad
\psi_m
=
\operatorname{GDF}(R_{\cdot m};\phi_{\psi}).
\label{eq:gdf}
\end{equation}

Here, $R_{i\cdot}$ and $R_{\cdot m}$ denote the response logs
of learner $u_i$ and item $I_m$, respectively. To incorporate
$Q$-matrix priors, the learner trait is obtained through residual
fusion,
$\theta_i=(1-\alpha)\theta_i^{\mathrm{imp}}
+\alpha\theta_i^{\mathrm{exp}}$,
where $\alpha\in[0,1]$ balances the implicitly learned
representation and the explicitly extracted $Q$-matrix
representation~\cite{li_gncdm_2026}.

\textbf{Aggregation Layer.}
Before decoding, concept-level traits are projected into a shared
embedding space via $Q$-gated aggregation. For learner $u_i$ and item $I_j$,
\begin{equation}
u_{ij}
=
W_{\theta}\left(\theta_i\odot Q_{j\cdot}\right)+b_{\theta},
\qquad
v_{ij}
=
W_{\psi}\left(\psi_j\odot Q_{j\cdot}\right)+b_{\psi}.
\label{eq:q-gated-aggregation}
\end{equation}
where $W_{\theta},W_{\psi}\in\mathbb{R}^{d\times T}$ and
$b_{\theta},b_{\psi}\in\mathbb{R}^{d}$. The $Q$-mask retains only the
concepts required by $I_j$. Under the monotonicity
assumption~~\cite{wang_ncdm_2020,li_gncdm_2026}, $W_{\theta}$ is
constrained to be non-negative (PosLinear), while $W_{\psi}$ remains a
standard linear map. The resulting embeddings $u_{ij}$ and $v_{ij}$ are
then fed into the item response function (IRF).

\textbf{Response Reconstruction (Item Response Function).}
The IRF predicts the probability of a correct response from the aggregated learner and item representations:

\begin{equation}
\hat{y}_{ij}
=
\operatorname{MLP}\left(u_{ij}-v_{ij}\right)
\in(0,1).
\label{eq:irf}
\end{equation}

where $u_{ij}-v_{ij}$ represents the gap between learner proficiency and
item demand, and the MLP produces the predicted response probability through
a sigmoid output layer. The model is trained on observed learner--item
interactions using binary cross-entropy loss~\cite{li_gncdm_2026}.

  \subsection{\textbf{Incremental Cognitive Diagnosis Setting}}
  Let $M_{\mathrm{old}}$ and $K_{\mathrm{old}}$ denote the numbers of
historical items and knowledge concepts, respectively. We train the
anchor model $f_{\Phi^{(0)}}$ on $\mathcal{D}_{\mathrm{old}}$ with an
$M_{\mathrm{old}}\times K_{\mathrm{old}}$ $Q$-matrix, yielding
$\theta_u^{(0)}\in\mathbb{R}^{K_{\mathrm{old}}}$.

In open instructional systems, the knowledge concept space and item bank
expand continuously as new knowledge concepts and items are introduced.
At incremental step $t$, the addition of $\Delta M$ new items and
$\Delta K$ new knowledge concepts expands the $Q$-matrix to
\(
\left(M_{\mathrm{old}}+\Delta M\right)
\times
\left(K_{\mathrm{old}}+\Delta K\right).
\)
The expanded model $f_{\Phi^{(t)}}$ must infer proficiency in the new
knowledge concepts and predict responses to the new items while
satisfying the representation-level stability constraint:
\begin{equation}
\theta_{u,[1:K_{\mathrm{old}}]}^{(t)}
=
\theta_u^{(0)},
\qquad
\forall u\in\mathcal{U}.
\label{eq:absolute-stability}
\end{equation}

We quantify deviations from this zero-drift condition using
Representation Drift (RD) in the following section.
  \subsection{\textbf{Quantitative Metric: Representation Drift}}
\textbf{Definition of RD.}
Representation Drift (RD) measures the Euclidean shift in the
representations of the same historical learners before and after
incremental learning. Analogous to semantic drift in class-incremental learning~\cite{yu_semantic_drift_2020}, RD quantifies changes in the historical representation space induced by incremental learning. In generative cognitive diagnosis, this space is defined by concept-aligned latent ability traits $\theta$, making RD a trait-space drift metric for the historical diagnostic subspace.

Let $\mathcal{U}_{\mathrm{old}}^{\mathrm{test}}$ denote the learners
evaluated on the old-item test set. For each learner
$i\in\mathcal{U}_{\mathrm{old}}^{\mathrm{test}}$, let
\(
\theta_i^{(0)}
\in
\mathbb{R}^{K_{\mathrm{old}}}
\quad\text{and}\quad
\theta_i^{(t)}
\in
\mathbb{R}^{K_{\mathrm{old}}+\Delta K}
\)
denote the latent traits inferred by the initial and incremental models,
respectively. The slice $[1:K_{\mathrm{old}}]$ selects the dimensions
corresponding to historical concepts. RD is
defined as
\begin{equation}
\mathrm{RD}
=
\frac{1}{
\left|\mathcal{U}_{\mathrm{old}}^{\mathrm{test}}\right|
\sqrt{K_{\mathrm{old}}}
}
\sum_{i\in\mathcal{U}_{\mathrm{old}}^{\mathrm{test}}}
\left\|
\theta_i^{(0)}
-
\theta_{i,[1:K_{\mathrm{old}}]}^{(t)}
\right\|_2.
\label{eq:representation-drift}
\end{equation}

When $\mathrm{RD}=0$, historical traits are pointwise invariant in the
old-concept subspace, indicating representation-level consistency. The
$\sqrt{K_{\mathrm{old}}}$ normalization reduces dimensionality effects
and enables comparison across models. However, RD only captures representation-level stability, while psychometric
consistency is reflected in unchanged old-item predictions under the
same historical inputs. Achieving the latter additionally relies on
Orthogonal Column Masking and Shared Bias Freezing, as shown in
Section~\ref{Ablation study}.

\section{Methodology}

\subsection{Overview of CLEAN}
CLEAN freezes the parameters associated with historical concepts and learns newly introduced concepts through isolated incremental branches. As illustrated in Figure~\ref{fig:clean-framework},
CLEAN integrates STB-based topological reordering, expandable Generative
Diagnosis Functions (GDFs), and an architecturally isolated aggregation
layer. Historical parameters remain unaffected by incremental gradients, while STB, Orthogonal Column Masking, and Shared Bias Freezing prevent new-concept updates from altering old-item diagnostic pathways. Consequently,
CLEAN preserves the historical trait subspace with $\mathrm{RD}=0$ and maintains unchanged old-item diagnostic outputs. Meanwhile, the trainable
capacity of the incremental branches is devoted to learning newly introduced
concepts. The following subsections present the STB protocol, the expandable CLEAN-GDF with micro-variance initialization and separate pathways for learner and item traits, and aggregation-layer isolation through Orthogonal Column Masking and Shared Bias Freezing.

\begin{figure}[t]
  \centering
  \includegraphics[width=0.8\linewidth]{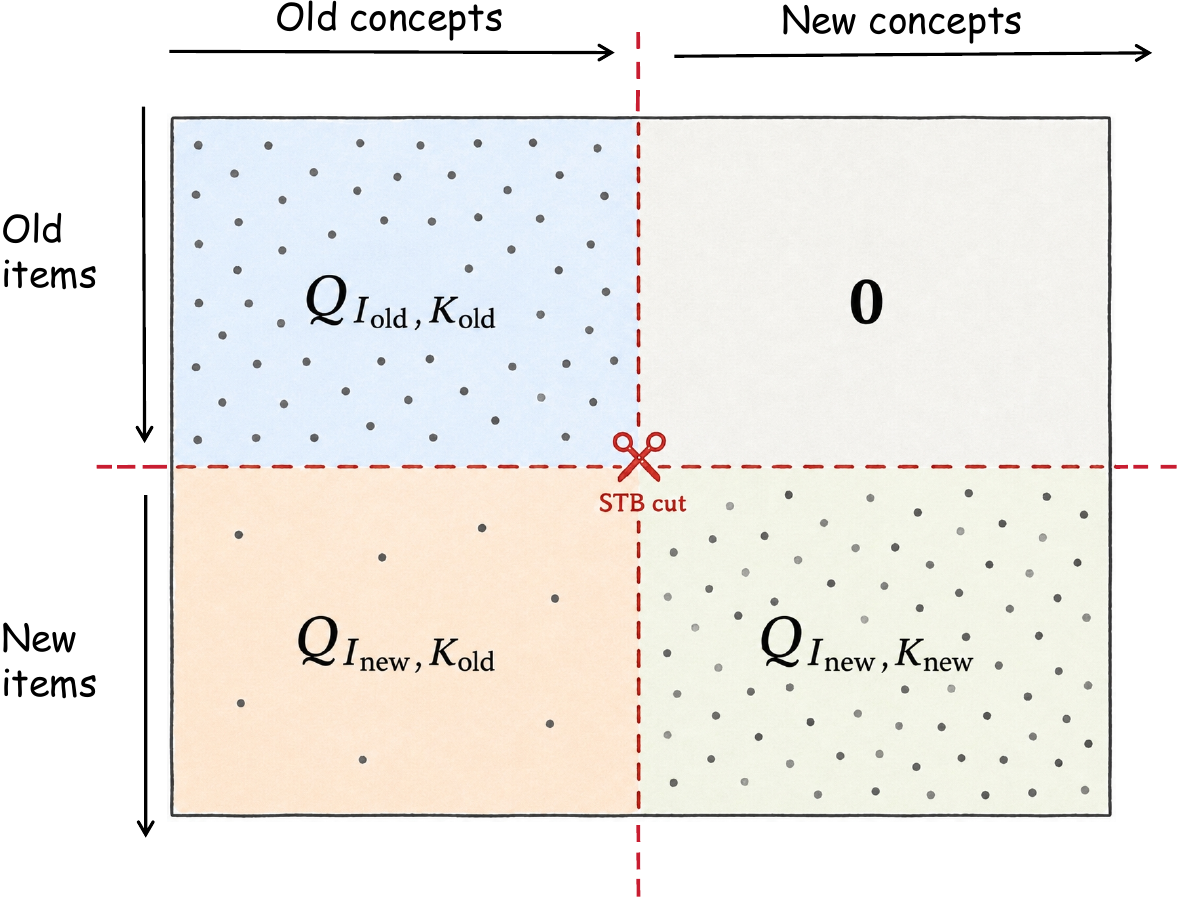}
  \caption{STB-Induced Block-Structured \(Q\)-Matrix.}
  \Description{The block-structured Q-matrix induced by STB, separating historical and newly introduced items and knowledge concepts.}
  \label{fig:stb-block-qmatrix}
\end{figure}
 \subsection{\textbf{STB Protocol}}
STB is a partitioning and reordering protocol operating on $Q$-matrix
indices. It establishes the topological prerequisites for the subsequent
structural partitioning of the computational graph. Given a new
knowledge concept set $\mathcal{K}_{\mathrm{new}}$, STB classifies items
based on whether they involve any new knowledge concepts.

Specifically, items involving at least one new knowledge concept are
classified as new items, whereas those involving only existing knowledge
concepts are classified as old items. After the item rows and concept
columns are reordered accordingly, the $Q$-matrix exhibits a clean
block-structured topology:
\[
Q=
\left[
\begin{matrix}
Q_{\mathcal{I}_{\mathrm{old}},\mathcal{K}_{\mathrm{old}}}
&
\mathbf{0}
\\
Q_{\mathcal{I}_{\mathrm{new}},\mathcal{K}_{\mathrm{old}}}
&
Q_{\mathcal{I}_{\mathrm{new}},\mathcal{K}_{\mathrm{new}}}
\end{matrix}
\right]
=
\left[
\begin{matrix}
Q_{11}
&
\mathbf{0}
\\
Q_{21}
&
Q_{22}
\end{matrix}
\right].
\]

There is a sound psychometric basis for setting the $Q$-matrix entries
between old items and new knowledge concepts to zero: a model trained
only on historical stages has never observed any signals related to the
new concepts. By enforcing that old items are not associated with new
knowledge concepts, the incremental evaluation benchmark remains
uncontaminated, ensuring that the forgetting and plasticity metrics for
historical items are not distorted by newly introduced concepts.

By contrast, traditional lateral-growth neural network
methods~\cite{rusu_pnn_2016} do not explicitly distinguish between
historical and newly added training data. As a result, the feature
information associated with the same knowledge concept may be
simultaneously encoded by parameters in both the old and new networks,
entangling historical and incremental diagnostic pathways. This
entanglement disrupts the historical diagnostic space and substantially
undermines measurement consistency.

The core mathematical property of STB lies in the upper-right block,
$Q_{12}\equiv\mathbf{0}$, indicating that historical items in
$\mathcal{I}_{\mathrm{old}}$ are not associated with any newly introduced
concepts in $\mathcal{K}_{\mathrm{new}}$. In contrast, new items in
$\mathcal{I}_{\mathrm{new}}$ may still involve historical concepts, so
$Q_{\mathcal{I}_{\mathrm{new}},\mathcal{K}_{\mathrm{old}}}$ can be
nonzero. This preserves the authentic topology in which newly introduced
items may assess both historical and new knowledge concepts.

The upper-right block topology, characterized by
$Q_{12}=\mathbf{0}$, provides the basis for the orthogonal column
masking described in Section~\ref{sec:Masking}, enabling simultaneous zeroing in
both the forward and backward paths. STB is a partitioning protocol
rather than a learnable model component; it defines the topological
boundaries between historical and newly added items and knowledge
concepts based on expert-annotated $Q$-matrices.

By definition, any item associated with at least one newly introduced
knowledge concept in $\mathcal{K}_{\mathrm{new}}$ is assigned to
$\mathcal{I}_{\mathrm{new}}$. Consequently, the items remaining in
$\mathcal{I}_{\mathrm{old}}$ naturally satisfy the condition
$Q_{12}=\mathbf{0}$.

 \subsection{\textbf{CLEAN-GDF}}
 As discussed earlier, cognitive diagnosis typically serves as an upstream task for downstream educational applications; therefore, the induced trait representations should remain as stable as possible across incremental updates. We need the predicted traits \(\theta\) and \(\psi\) themselves to evolve smoothly. To this end, we adapt the GDF via a micro-variance initialization scheme for newly grown branches.

Instead of standard Xavier initialization~\cite{glorot_xavier_2010}, the newly grown network weights are initialized from a truncated Gaussian distribution scaled down by a factor of \(10^{-3}\) relative to the variance of the existing parameters. This prevents abrupt disturbances to the established diagnostic space while allowing the new parameters to receive gradients and gradually learn new concepts.

\begin{equation}
W_{\mathrm{new}}^{(a,b)}
\sim
\mathcal{N}_{[-2\sigma,\,2\sigma]}
\left(0,\sigma^2\right),
\qquad
\sigma
=
\sqrt{\operatorname{Var}\left(W_{\mathrm{ref}}\right)\cdot 10^{-3}}.
\label{eq:micro_variance_initialization}
\end{equation}

where each newly grown parameter \(W_{\mathrm{new}}^{(a,b)}\) is sampled from a zero-mean Gaussian distribution truncated to the interval \([-2\sigma,2\sigma]\).

First, to avoid ``zero-gradient death,'' new branches are initialized with small nonzero weights. If the weights of a new branch are initialized to zero, both its forward activations and backward gradients become zero, preventing the branch from learning. This issue is particularly severe for LoRA variant~\cite{hu_lora_2022}
, where setting both factors in
\(\Delta W=BA\) to zero causes their gradients to vanish. We therefore
initialize
\(A,B\sim\mathcal{N}(0,10^{-6})\),
enabling both factors to update from the first mini-batch while keeping
the initial contribution of \(\Delta W=BA\) near zero.

Second, micro-variance initialization places the newly added full-rank branch in a near-silent initial state: new weights are drawn from a truncated Gaussian whose variance is \(10^{-3}\) times that of the frozen reference parameters. Consequently, training starts from a well-conditioned baseline close to the anchor point and gradually accumulates capacity, avoiding large random perturbations at the beginning and leading to more stable convergence.

Importantly, although $\theta$ and $\psi$ follow the same
micro-variance growth principle, they use separate expansion pathways
rather than sharing a common incremental trunk. Learner traits $\theta$
are generated by the response-to-ability encoder $f(\cdot)$ and its new
branch $f_{\mathrm{new}}$, whereas item traits $\psi$ are generated by
the dual encoder $g(\cdot)$ and its new branch $g_{\mathrm{new}}$.

For learner traits, each response vector is partitioned into old- and
new-item blocks.
The frozen encoder $f_{\mathrm{old}}$ processes old items to produce
$\theta_{\mathrm{old}}$, while $f_{\mathrm{new}}$ processes
new items to produce $\theta_{\mathrm{new}}$. The two
outputs are then concatenated along the concept dimension. For item
traits, the complete learner-wise response column is fed into both
pathways without input partitioning. Likewise, the frozen encoder $g_{\mathrm{old}}$ produces
$\psi_{\mathrm{old}}$, while $g_{\mathrm{new}}$ produces
$\psi_{\mathrm{new}}$. The two representations
are then concatenated along the concept dimension and passed to the
subsequent aggregation layer.

This asymmetric design reflects the different structures of learner and
item response logs. A learner response vector contains interactions with
both old and new items, so input partitioning is used to prevent gradients from new items from disturbing historical ability dimensions. In contrast, each item is treated as a single entity, and its diagnostic input is already a complete user-wise response column. Accordingly, the $\psi$ pathway does not exhibit the cross-block
interference observed in the learner pathway and therefore does not
require the response column to be partitioned into old- and new-item
blocks.

 \subsection{\textbf{CLEAN Aggregation Layer}}
The aggregation layer projects the concept-level traits \(\theta,\psi\in\mathbb{R}^{K}\) into dense representations in \(\mathbb{R}^{d}\), which are subsequently fed into the IRF. After topological expansion, the old and newly introduced concept traits are concatenated along the concept dimension. The expanded concept representation is then projected by the same aggregation layer into a fixed-dimensional dense representation for the IRF. If the historical weight columns or the shared bias were allowed to change, even a frozen GDF would no longer reproduce the anchored traits of old items. Therefore, the aggregation layer must simultaneously support plasticity for newly introduced concepts while preserving the historical aggregation subspace in a bitwise-identical state with respect to the foundation-stage anchor. We achieve this through two complementary mechanisms: Deterministic Orthogonal Column Masking for the concept-wise weights (Section~\ref{sec:Masking}) and Shared Bias Freezing for the aggregation bias (Section~\ref{sec:Bias}).

\subsubsection{Deterministic Orthogonal Column Masking.}
\label{sec:Masking}
This mechanism is central to achieving pointwise invariance of historical diagnostic outputs. Rather than explicitly applying a binary vector, the mask is implicitly induced at the aggregation layer by the STB structure with \(Q_{12}=0\) through two steps: chunk-wise slicing and incremental block de-biasing. For a given item \(I_m\), aggregation is decoupled across the historical and new subspaces:

\begin{equation}
\operatorname{agg}(m)
=
\underbrace{
W^{\mathrm{old}}
\left(
\theta_{\mathrm{old}}\odot q_m^{\mathrm{old}}
\right)
+b^{\mathrm{old}}
}_{\text{frozen historical block with bias}}
+
\underbrace{
W^{\mathrm{new}}
\left(
\theta_{\mathrm{new}}\odot q_m^{\mathrm{new}}
\right).
}_{\text{incremental block without bias}}
\label{eq:decoupled_aggregation}
\end{equation}

For any historical item \(I_m\in\mathcal{I}_{\mathrm{old}}\), STB ensures that the columns corresponding to new knowledge concepts are zero, yielding \(\theta_{\mathrm{new}}\odot q_m^{\mathrm{new}}=0\). Structurally, the forward prediction for old items contains no contribution from new concepts. Because the historical aggregation columns \(W^{\mathrm{old}}\) and the shared bias \(b\) (Section~\ref{sec:Bias}) are frozen at their anchor values, old-item predictions remain identical to those of the anchor model.

This mask is termed orthogonal because it effectively decouples parameter updates between the historical and newly introduced subspaces along the concept dimension. To clarify this property, we distinguish two complementary separation mechanisms. First, the clean forward pass for historical items is guaranteed by STB, where the new-concept columns are zero for historical items. As a result, the new-concept components are always multiplied by zero, independently of the column mask. Second, the column mask operates during the backward pass: it zeroes out the gradients of the historical aggregation weight columns, i.e., \(\bigl[:,\, :K_{\mathrm{old}}\bigr]\), preventing them from drifting when training on new items that may still involve historical concepts. In this way, the weights of the historical concept columns remain locked at their anchor points. It should also be emphasized that the column mask applies only to weights, not biases. The stability of the shared bias in the aggregation layer is guaranteed by the bias-freezing mechanism described in Section~\ref{sec:Bias}. Furthermore, during the forward pass, the incremental diagnostic branch uses only responses associated with new items, denoted as \(\mathbf{x}_{\mathrm{new}}=\texttt{user\_log}[:,\,I_{\mathrm{old}}:]\), to generate \(\theta_{\mathrm{new}}\), instead of reusing hidden-layer features from the historical subnet. This design cuts off the coupling chain at its source and ensures that the gradient graph of the frozen historical network remains completely disjoint from that of the incremental branch.

\subsubsection{Parameter Bias Freezing.}
\label{sec:Bias}
The aggregation layer contains a shared bias term \(b\) in addition to the weight columns corresponding to historical and new concepts. Since the orthogonal column mask applies only to weights, the bias term \(b\) would still be updated by the optimizer during incremental training if left unconstrained. Therefore, a second layer of isolation is required. To this end, we exclude the shared bias of the aggregation layer from the optimizer and freeze it at the anchor value \(b^{old}\), restricting incremental updates to the new-concept branch and the weight columns associated with new concepts. Consequently, for any historical input, the aggregated output remains \(W^{\mathrm{old}}(\theta_{\mathrm{old}}\odot q_{\mathrm{old}})+b^{old}\), ensuring bitwise identity with the anchor state. Bias freezing and orthogonal column masking address complementary aspects of architectural isolation and are both indispensable. Removing either mechanism immediately degrades prediction performance on old items, as shown in Section~\ref{Ablation study} and Table~\ref{tab:ablation-results}. Together, these mechanisms lock the entire historical aggregation subspace.

\begin{table*}[!t]
\centering
\caption{
Performance comparison on the Junyi, a0910, and Math1 datasets under
the Random-Split setting. ↑ (↓) means the higher (lower) score the better performance.
}
\label{tab:main_results}

\scriptsize
\setlength{\tabcolsep}{2.5pt}
\renewcommand{\arraystretch}{1.42}
\setlength{\extrarowheight}{0.7pt}

\resizebox{\textwidth}{!}{%
\begin{tabular}{@{}ll*{12}{c}@{}}
\toprule

\multirow{2}{*}{\textbf{Paradigm}}
&
\multirow{2}{*}{\textbf{Model}}
&
\multicolumn{4}{c}{\textbf{Junyi}}
&
\multicolumn{4}{c}{\textbf{ASSIST}}
&
\multicolumn{4}{c}{\textbf{Math1}}
\\

\cmidrule(lr){3-6}
\cmidrule(lr){7-10}
\cmidrule(lr){11-14}

&
&
\makecell{\textbf{ACC$_{\mathrm{old}}$}\\$\uparrow$}
&
\makecell{\textbf{ACC$_{\mathrm{new}}$}\\$\uparrow$}
&
\makecell{\textbf{ACC$_{\mathrm{all}}$}\\$\uparrow$}
&
\makecell{\textbf{RD}\\$\downarrow$}
&
\makecell{\textbf{ACC$_{\mathrm{old}}$}\\$\uparrow$}
&
\makecell{\textbf{ACC$_{\mathrm{new}}$}\\$\uparrow$}
&
\makecell{\textbf{ACC$_{\mathrm{all}}$}\\$\uparrow$}
&
\makecell{\textbf{RD}\\$\downarrow$}
&
\makecell{\textbf{ACC$_{\mathrm{old}}$}\\$\uparrow$}
&
\makecell{\textbf{ACC$_{\mathrm{new}}$}\\$\uparrow$}
&
\makecell{\textbf{ACC$_{\mathrm{all}}$}\\$\uparrow$}
&
\makecell{\textbf{RD}\\$\downarrow$}
\\

\specialrule{0.55pt}{3pt}{3pt}

% =========================================================
% Anchor
% =========================================================

Anchor
&
G-NCDM
&
$\second{78.40\pm0.15}$
&
\NA
&
$78.40\pm0.15$
&
\NA
&
$\second{73.51\pm0.15}$
&
\NA
&
$73.51\pm0.15$
&
\NA
&
$72.14\pm0.51$
&
\NA
&
$72.14\pm0.51$
&
\NA
\\

\specialrule{0.45pt}{3pt}{3pt}

% =========================================================
% Ours
% =========================================================

\multirow{2}{*}{Ours}
&
CLEAN-Full
&
${\second{78.40\pm0.15}}^{*}$
&
${\second{74.53\pm0.30}}^{*}$
&
${\second{77.19\pm0.11}}^{*}$
&
$\best{0.00\pm0.00}$
&
${\second{73.51\pm0.15}}^{*}$
&
${\second{72.21\pm0.19}}^{*}$
&
${\second{73.11\pm0.15}}^{*}$
&
$\best{0.00\pm0.00}$
&
$72.14\pm0.51$
&
$72.25\pm4.69$
&
$72.18\pm1.62$
&
$\best{0.00\pm0.00}$
\\

&
CLEAN-LoRA
&
${\second{78.40\pm0.15}}^{*}$
&
$73.57\pm0.64$
&
$76.89\pm0.19$
&
$\best{0.00\pm0.00}$
&
${\second{73.51\pm0.15}}^{*}$
&
${72.12\pm0.24}^{*}$
&
$73.08\pm0.17$
&
$\best{0.00\pm0.00}$
&
$72.14\pm0.51$
&
${68.50\pm3.00}$
&
$70.85\pm1.01$
&
$\best{0.00\pm0.00}$
\\

\specialrule{0.45pt}{3pt}{3pt}
% =========================================================
% Baselines
% =========================================================

\multirow{6}{*}{Baselines}
&
Full-Replay Oracle
&
$\best{78.55\pm0.25}$
&
$\best{75.47\pm0.22}$
&
$\best{77.58\pm0.18}$
&
$4.88\pm1.29$
&
$\best{73.66\pm0.12}$
&
$\best{72.99\pm0.33}$
&
$\best{73.45\pm0.17}$
&
$6.92\pm0.72$
&
$\second{72.20\pm0.28}$
&
$\second{75.12\pm0.28}$
&
$\second{73.23\pm0.17}$
&
$6.47\pm1.60$
\\

&
EWC
&
$71.54\pm2.03$
&
$71.58\pm0.40$
&
$71.56\pm1.41$
&
$8.92\pm0.20$
&
$66.87\pm0.67$
&
$64.09\pm0.61$
&
$66.01\pm0.62$
&
$9.01\pm0.09$
&
$68.59\pm0.51$
&
$73.27\pm0.86$
&
$70.24\pm0.53$
&
$9.70\pm0.26$
\\

&
DER++
&
$76.53\pm0.28$
&
$74.27\pm0.26$
&
$75.82\pm0.22$
&
$\second{3.09\pm1.14}$
&
$69.50\pm0.44$
&
$67.66\pm0.51$
&
$68.93\pm0.28$
&
$\second{3.64\pm0.94}$
&
$71.72\pm0.52$
&
$\best{75.52\pm0.22}$
&
$73.06\pm0.33$
&
$\second{1.32\pm1.11}$
\\

&
C-LoRA
&
$71.26\pm2.01$
&
$73.12\pm0.36$
&
$71.84\pm1.34$
&
$19.69\pm0.27$
&
$67.64\pm0.28$
&
$64.28\pm0.46$
&
$66.61\pm0.23$
&
$13.11\pm0.09$
&
$68.88\pm0.51$
&
$71.98\pm0.68$
&
$69.97\pm0.46$
&
$10.82\pm0.27$
\\

&
X-DER
&
$77.59\pm0.42$
&
$73.19\pm0.69$
&
$76.21\pm0.23$
&
$6.36\pm1.06$
&
$72.86\pm0.24$
&
$69.83\pm0.44$
&
$71.92\pm0.10$
&
$7.32\pm1.57$
&
$\best{72.54\pm0.21}$
&
$\second{75.12\pm0.35}$
&
$\best{73.45\pm0.16}$
&
$7.24\pm2.00$
\\

&
ICD
&
$70.01\pm1.47$
&
$66.84\pm4.50$
&
$69.02\pm0.40$
&
$119.02\pm24.52$
&
$54.40\pm0.00$
&
$65.00\pm0.00$
&
$57.67\pm0.00$
&
$422.42\pm0.00$
&
$68.62\pm0.00$
&
$71.10\pm0.00$
&
$69.49\pm0.00$
&
$95.75\pm0.00$
\\

\bottomrule
\end{tabular}%
}

\vspace{4pt}

\begin{minipage}{0.99\linewidth}
\footnotesize
\raggedright
\textit{Note.}
Full-Replay Oracle uses all historical training data and full-parameter
expansion, 
serving as a theoretical performance upper bound. The anchor model is evaluated only on the old-item test set,  "--" indicates that the corresponding result is unavailable.
Statistical significance is assessed using paired $t$-tests over
10 independent runs.Bold and underlined values indicate the best and second-best results, respectively. 
The superscript $^{*}$ indicates that the corresponding CLEAN result is
significantly better than all compared non-oracle baselines at $p<0.05$.

\end{minipage}

\end{table*}

\section{ \textbf{EXPERIMENTS}
}
 \subsection{\textbf{Experimental Setup}}

We use three real-world datasets: Junyi, ASSISTments 2009--2010 Skill Builder (ASSIST), and Math1. Junyi is derived from Junyi Academy by selecting the 1,000 most active students, yielding a $712 \times 39$ $Q$-matrix with 71.36\% response sparsity, which is used
to evaluate model performance under normal interaction conditions. ASSIST,
based on the ASSISTments online tutoring system, contains 4,163 students and 17,746 exercises after preprocessing, with a $17{,}746 \times 123$ $Q$-matrix and 99.64\% sparsity, which is used to evaluate model performance under extremely
data-scarce scenarios. Math1, a classic cognitive diagnosis benchmark,
contains 4,209 students, 20 exercises, and 11 knowledge concepts, yielding a $20 \times 11$ $Q$-matrix. Its response matrix is fully observed, comprising 84,180 interactions with 0.00\% sparsity, which is used to
evaluate model performance under extremely interaction-dense scenarios.

\noindent\textbf{Data Split.}
We use a 70\%/10\%/20\% train/validation/test split. Under the
random-split setting, learner--item interactions are randomly partitioned
while allowing the same learners to appear in all three sets. This setting
isolates incremental item- and concept-space expansion from cold-start
learner adaptation. An additional user-split evaluation with mutually
exclusive learner groups is reported in Appendix~\ref{app:user-split}.

\textbf{\textbf{Experimental Process}}\textbf{\textbf{:}} All methods follow the standard two-stage training protocol commonly used in continual learning. Based on the Q-matrix, items are divided into mutually exclusive old and new sets. Old items involve only previously learned knowledge concepts, whereas new items may involve both old and newly introduced concepts, thereby establishing clear boundaries for incremental learning. In Phase~1 (\textit{Foundation Training}), the model is trained on old items within the historical item--concept space
$\mathcal{I}_{\mathrm{old}} \times \mathcal{K}_{\mathrm{old}}$. In Phase~2  (\textit{Incremental Training}), the model is initialized from the foundation model and
incrementally trained on new items after expanding to
$\left(\mathcal{I}_{\mathrm{old}} \cup \Delta\mathcal{I}\right)
\times
\left(\mathcal{K}_{\mathrm{old}} \cup \Delta\mathcal{K}\right)$. After incremental training, each model is evaluated once on separate old-item and new-item test sets. We use the checkpoint obtained at convergence for each of the ten random seeds and report the average performance across all ten runs. Old-item metrics measure retention after incremental learning, whereas new-item metrics measure adaptation to newly introduced items. In addition, G-NCDM serves as the anchor baseline~\cite{li_gncdm_2026} and is evaluated only on the old-item test set.

\textbf{\textbf{Baselines}}:We compare CLEAN with six representative baselines: Full-Replay Oracle, EWC, DER++, C-LoRA, X-DER, and ICD~\cite{tong_icd_2022,kirkpatrick_ewc_2017,buzzega_derpp_2020,boschini_xder_2023,zhang_clora_2025}. These methods are divided into two groups. \textbf{1) Same-Backbone Comparison:} Full-Replay Oracle, X-DER, C-LoRA, and CLEAN are implemented on the same G-NCDM backbone, enabling direct comparison of RD. X-DER is architecture-agnostic and can be faithfully applied to G-NCDM through sample replay, while C-LoRA performs low-rank updates on the shared backbone with the pre-trained parameters frozen. \textbf{2) Cross-Backbone Reference:} EWC and DER++ assume fixed parameter dimensions and output spaces, whereas ICD was designed for its own diagnostic architecture. Adapting them to the expandable G-NCDM backbone would require additional structural modifications not included in the original methods. Therefore, EWC, DER++, and ICD are evaluated under their original implementations as cross-backbone references. 

\textbf{Evaluation Metrics:}
Following prior studies~\cite{tong_irr_2021,wang_ncdm_2020},
we evaluate model performance indirectly through
response prediction. We report ACC, F1, and
RMSE separately on old and new items using the subscripts
$\mathrm{old}$ and $\mathrm{new}$. Higher ACC and F1, and lower RMSE
indicate better performance. We also report
$\mathrm{ACC}_{\mathrm{all}}$
over all test interactions and RD for historical representation stability.
Unless otherwise stated, ACC is used as the primary metric for
statistical comparisons. Detailed metric results
for each dataset are reported in Appendix~\ref{app:detailed-metrics}.

\textbf{\textbf{Implementation Details:}} We initialize the base-network weights using Xavier
initialization~\cite{glorot_xavier_2010} and the newly expanded CLEAN
branches using micro-variance initialization with a scaling factor of
$10^{-3}$. All models are optimized using Adam~\cite{kingma_adam_2015}
with a learning rate of $10^{-3}$ and the binary cross-entropy loss.
Each model is trained for 25 epochs under Random-Split and User-Split. The residual mixing coefficient $\alpha$ is selected
separately for each dataset and split. Under Random-Split, $\alpha$ is
set to 0.20 for Math1 and 0.10 for both a0910 and Junyi. For a fair comparison, the hyperparameters of all baselines are selected
based on the highest validation
$\mathrm{ACC}_{\mathrm{all}}$
. All models are implemented in
PyTorch, and all experiments are conducted on a Linux server equipped
with a single NVIDIA H100 GPU with 80~GB of memory, 22 CPU cores, and
240~GB of RAM.
An anonymized implementation is available at
\url{https://anonymous.4open.science/r/CLEAN-2E32/}.
% Force the table to start at the top of a new page.
\begin{figure}[t]
  \centering
  \includegraphics[width=\columnwidth]{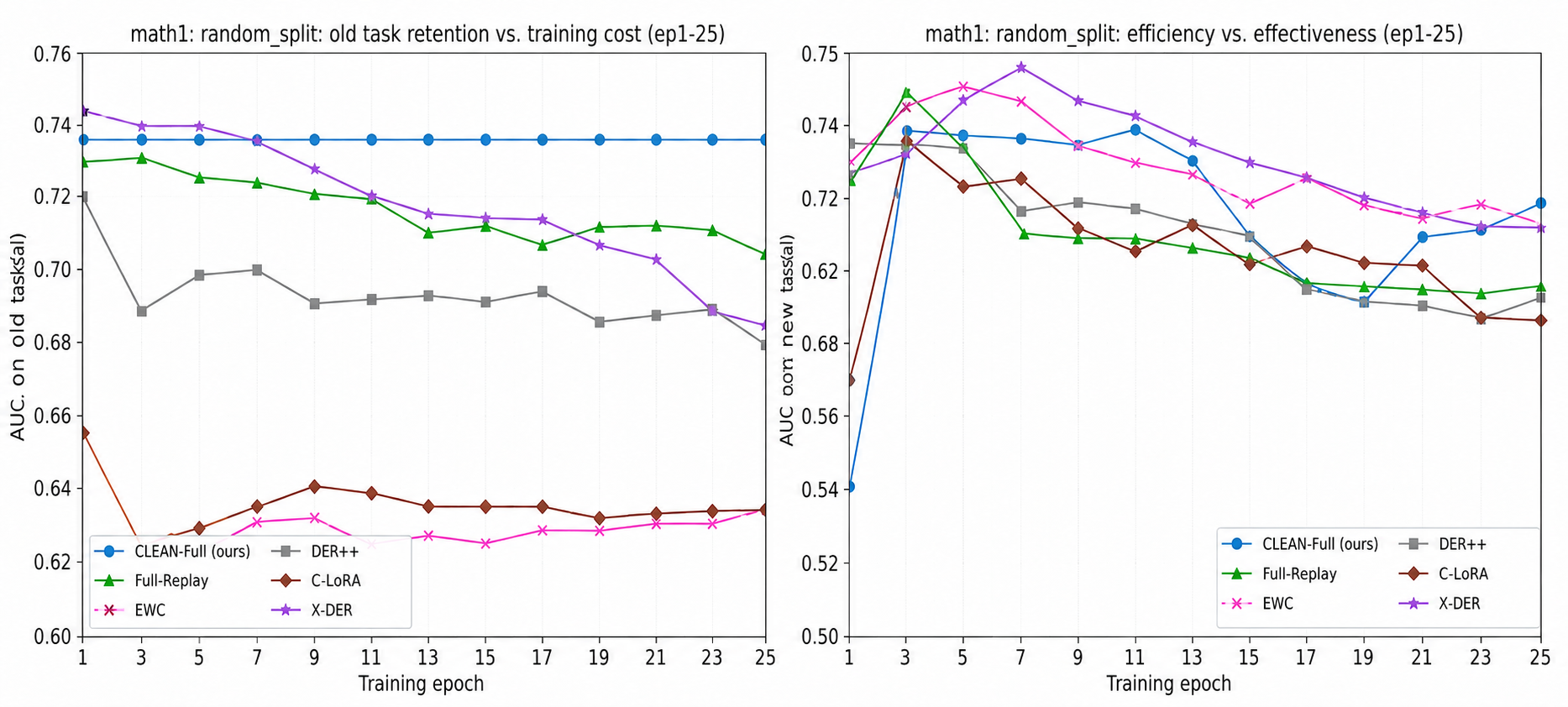}
 \caption{Training dynamics of
$\mathrm{ACC}_{\mathrm{new}}$ and
$\mathrm{ACC}_{\mathrm{old}}$ on Math1 under Random-Split.}
\Description{Training curves of old-item and new-item accuracy on
Math1 under the Random-Split setting.}
\label{fig:training-dynamics}
\end{figure}

\FloatBarrier
 \subsection{\textbf{Experimental Results}}

The Random-Split setting ensures that each learner appears in the training, validation, and test sets. We adopt it as the primary setting because it provides a controlled evaluation of incremental prediction: the model must accommodate newly added items and knowledge concepts while continuing to predict responses for previously observed learners.

\textbf{Overall Comparison.} Table~\ref{tab:main_results} reports $\mathrm{ACC}_{\mathrm{old}}$, $\mathrm{ACC}_{\mathrm{new}}$, $\mathrm{ACC}_{\mathrm{all}}$, and Representation Drift (RD) on Junyi, ASSISTments a0910, and Math1. G-NCDM denotes the foundation-stage anchor trained only on old items. Across all three datasets, CLEAN-Full and CLEAN-LoRA exactly match the anchor in $\mathrm{ACC}_{\mathrm{old}}$ and consistently achieve $\mathrm{RD}=0$, demonstrating that architectural isolation preserves the historical diagnostic subspace during incremental learning. On Junyi, CLEAN-Full achieves an $\mathrm{ACC}_{\mathrm{new}}$ of 74.53 and an $\mathrm{ACC}_{\mathrm{all}}$ of 77.19, ranking second only to Full-Replay Oracle, which obtains 75.47 and 77.58, respectively. CLEAN-Full also outperforms all non-oracle baselines in overall accuracy. On ASSISTments a0910, CLEAN-Full achieves an $\mathrm{ACC}_{\mathrm{new}}$ of 72.21 and an $\mathrm{ACC}_{\mathrm{all}}$ of 73.11,
again ranking second only to Full-Replay Oracle and outperforming the remaining continual-learning baselines. On Math1, CLEAN-Full obtains an $\mathrm{ACC}_{\mathrm{new}}$ of 72.25 while exactly retaining the anchor-level $\mathrm{ACC}_{\mathrm{old}}$ of 72.14. Although several baselines achieve higher new-item accuracy on Math1, CLEAN-Full provides a better balance between preserving historical representations and learning newly introduced items.

\begin{figure}[t]
  \centering
  \includegraphics[width=\columnwidth]
  {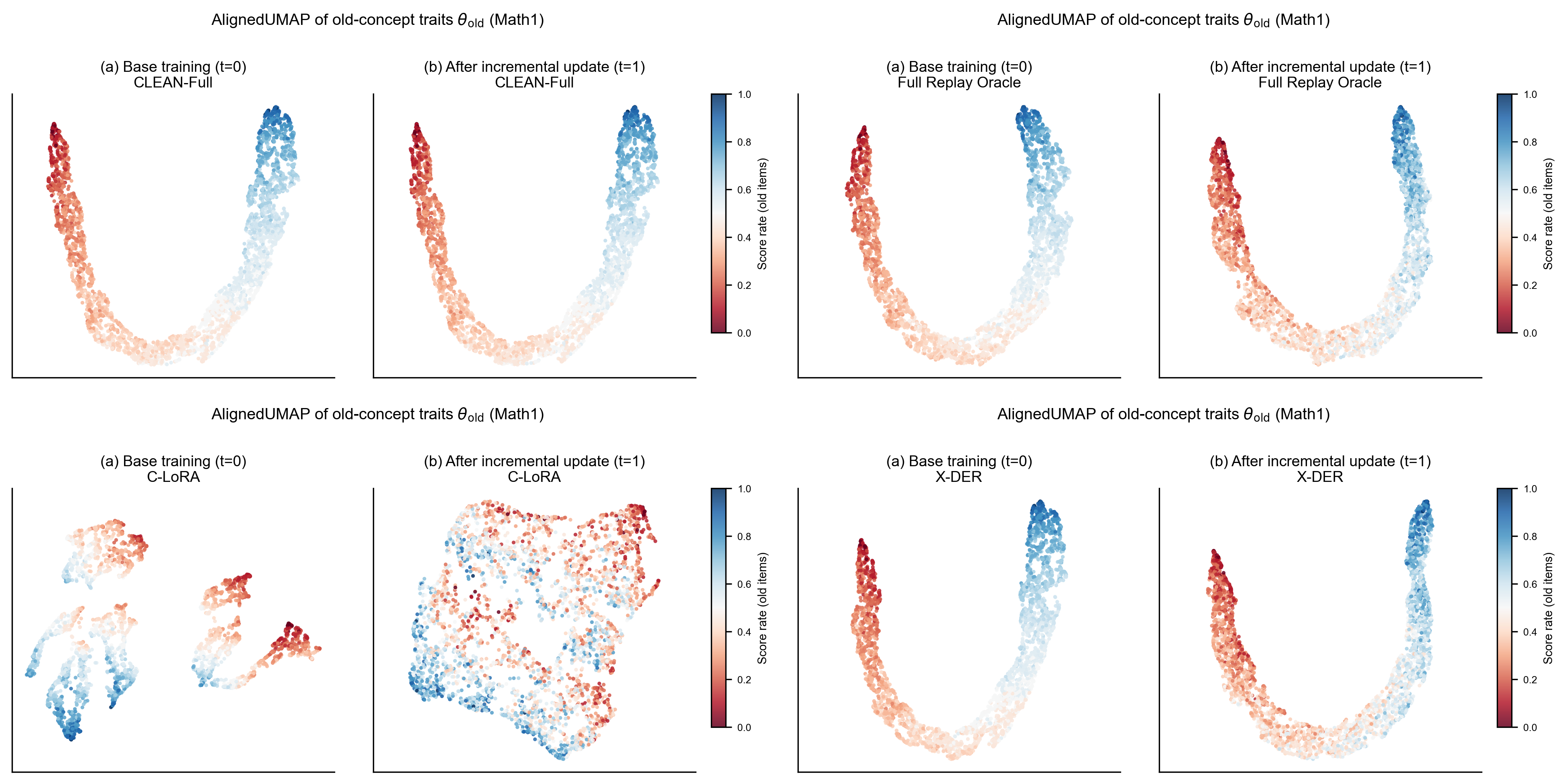}
  \caption{UMAP visualization of historical representation drift for four methods on the Math1 dataset. The left and right sides show the ends of the foundation and incremental training phases, respectively.}
  \Description{UMAP visualization comparing the historical representations before and after incremental learning for four methods on the Math1 dataset.}
  \label{fig:math1-umap-old-drift}
\end{figure}

\textbf{Psychometric Consistency.} Figure~\ref{fig:math1-umap-old-drift} visualizes the AlignedUMAP~\cite{dadu_aligned_umap_2023} projections of the old-concept trait subspace $\boldsymbol{\theta}_{\mathrm{old}}$ on Math1 before ($t=0$) and after ($t=1$) incremental learning. All four visualized methods use the same G-NCDM backbone, and each point is colored according to the learner’s observed accuracy on old items. For CLEAN-Full, the two embeddings perfectly overlap, preserving both the global U-shaped geometry and the accuracy gradient. This visual stability is consistent with its $\mathrm{RD}=0.00$ in Table~\ref{tab:main_results}. In contrast, Full-Replay Oracle and X-DER retain the general U-shaped structure but exhibit noticeable displacement after incremental learning. C-LoRA shows the greatest deformation, with initially fragmented clusters becoming diffuse and irregular after the update. These observations are consistent with their nonzero RD values. Among the four methods, only CLEAN-Full shows no observable displacement. These results show that CLEAN preserves the historical representation
space through architectural isolation, with $\mathrm{RD}=0$ serving as
a representation-level indicator of psychometric consistency. In the
following subsection, the ablation study further indicates that
Orthogonal Column Masking and Shared Bias Freezing are both required
to retain anchor-level performance on old items in this setting.

\begin{table}[!t]
  \centering
  \caption{Ablation study results.}
  \label{tab:ablation-results}

  \scriptsize
  \setlength{\tabcolsep}{1.5pt}
  \renewcommand{\arraystretch}{1.18}

  \resizebox{\columnwidth}{!}{%
  \begin{tabular}{@{}lccccccc@{}}
    \toprule
    \textbf{Architecture}
    & \makecell{\textbf{ACC}\\$_{\mathrm{old}}\uparrow$}
    & \makecell{\textbf{ACC}\\$_{\mathrm{new}}\uparrow$}
    & \makecell{\textbf{F1}\\$_{\mathrm{old}}\uparrow$}
    & \makecell{\textbf{F1}\\$_{\mathrm{new}}\uparrow$}
    & \makecell{\textbf{RMSE}\\$_{\mathrm{old}}\downarrow$}
    & \makecell{\textbf{RMSE}\\$_{\mathrm{new}}\downarrow$}
    & \makecell{\textbf{RD}\\$\downarrow$} \\
    \midrule

    Anchor
    & \textbf{69.42}
    & --
    & \textbf{78.69}
    & --
    & \textbf{45.64}
    & --
    & -- \\

    CLEAN w/o OCM and SBF
    & 44.46
    & \textbf{69.16}
    & 38.06
    & 76.94
    & 54.73
    & \underline{45.15}
    & \textbf{0.00} \\

    CLEAN w/o OCM
    & \underline{60.22}
    & 68.50
    & \underline{67.63}
    & \textbf{78.14}
    & \underline{48.78}
    & \textbf{45.07}
    & \textbf{0.00} \\

    CLEAN w/o SBF
    & 55.99
    & \underline{68.91}
    & 60.29
    & \underline{78.09}
    & 49.53
    & 45.26
    & \textbf{0.00} \\

    \textbf{CLEAN-Full}
    & \textbf{69.42}
    & 67.52
    & \textbf{78.69}
    & 77.91
    & \textbf{45.64}
    & 45.32
    & \textbf{0.00} \\

    \bottomrule
  \end{tabular}%
  }

  \vspace{2pt}

  \begin{minipage}{\columnwidth}
    \scriptsize
    \raggedright
    \textit{Note.}
    OCM denotes Orthogonal Column Masking, and SBF denotes Shared Bias
    Freezing.  ``--'' indicates that the corresponding result is unavailable.
  \end{minipage}
\end{table}

\vspace{2pt}

\begin{figure}[t]
  \centering
  \includegraphics[width=1\columnwidth]
  {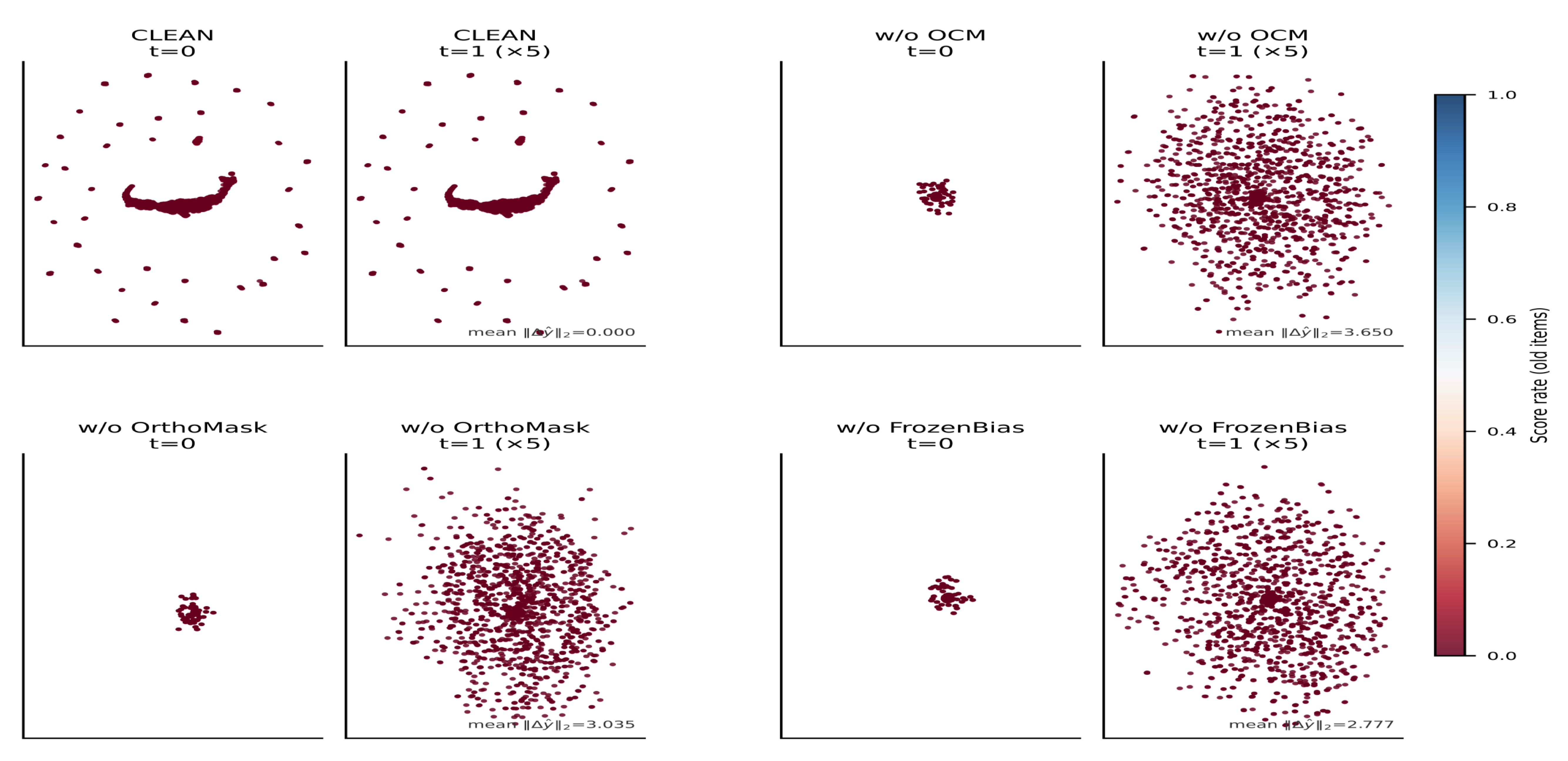}
  \caption{
  UMAP visualization of prediction-space drift for different ablation
  variants on the a0910 dataset. Each pair shows learner embeddings in the old-item prediction space at $t{=}0$ (left) and after incremental training at $t{=}1$ (right; $\times5$ for display). Points are colored by score rate on old items.
  }
  \Description{
  UMAP visualization comparing prediction representations before and
  after incremental learning for different CLEAN ablation variants on
  the a0910 dataset.
  }
  \label{fig:a0910-umap-ablation}
\end{figure}

\textbf{Takeaway.} Under the Random-Split setting, CLEAN achieves exact retention of the historical trait subspace, with $\mathrm{RD}=0$ and anchor-level $\mathrm{ACC}_{\mathrm{old}}$ across all three datasets. CLEAN-Full ranks second only to Full-Replay Oracle on Junyi and ASSISTments a0910, while maintaining a competitive balance between stability and plasticity on Math1.

 \subsection{\textbf{Ablation study}}
 \label{Ablation study}
The conventional continual learning baselines in Table~\ref{tab:main_results} alleviate forgetting but still exhibit nonzero Representation Drift ($\mathrm{RD}>0$). One way to achieve 
$\mathrm{RD}=0$ is PNN-style architectural isolation~\cite{rusu_pnn_2016}, which freezes the historical subnet and adds a lateral branch for new concepts. In CLEAN, this setting corresponds to \textbf{CLEAN w/o OCM and SBF}, which retains the lateral GDF expansion but removes both Orthogonal Column Masking and Shared Bias Freezing. As a result, the historical aggregation weights and shared bias remain trainable, allowing incremental gradients to alter downstream representations. Based on the same backbone, we further evaluate the individual contributions of Orthogonal Column Masking and Shared Bias Freezing on ASSISTments 2009--2010 under the random-split setting, as reported in Table~\ref{tab:ablation-results}.

All ablation variants achieve $\mathrm{RD}=0$, showing that freezing the historical GDF is sufficient to preserve the historical trait subspace measured by RD.  However, zero representation drift does not necessarily guarantee unchanged predictions on old items because the shared aggregation layer remains downstream of the frozen GDF. Removing Orthogonal Column Masking reduces $\mathrm{ACC}_{\mathrm{old}}$ from 69.42 to 60.22 and $\mathrm{F1}_{\mathrm{old}}$ from 78.69 to 67.63, while increasing $\mathrm{RMSE}_{\mathrm{old}}$ from 45.64 to 48.78. Removing Shared Bias Freezing causes a larger decline, with $\mathrm{ACC}_{\mathrm{old}}$ falling to 55.99, $\mathrm{F1}_{\mathrm{old}}$ to 60.29, and $\mathrm{RMSE}_{\mathrm{old}}$ rising to 49.53. When both mechanisms are removed in CLEAN w/o OCM, the degradation is most severe: $\mathrm{ACC}_{\mathrm{old}}$ drops to 44.46, $\mathrm{F1}_{\mathrm{old}}$ to 38.06, and $\mathrm{RMSE}_{\mathrm{old}}$ increases to 54.73. By contrast, the new-item metrics vary only moderately across the variants, indicating that these two mechanisms primarily protect historical predictive pathways rather than improve adaptation to new items. Therefore, freezing the GDF alone is insufficient, and preserving the historical aggregation subspace requires both Orthogonal Column Masking and Shared Bias Freezing. Because RD measures only the GDF-level historical trait subspace, it cannot capture downstream prediction drift. To visualize such prediction-level drift, Figure~\ref{fig:a0910-umap-ablation} presents the AlignedUMAP embeddings of the old-item prediction vectors $\mathbf{y}_{\mathrm{old}}$ before and after incremental learning.

\section{ \textbf{CONCLUSION}}

In this study, we introduced CLEAN, an incremental cognitive diagnosis framework that encodes psychometric consistency as a hard architectural constraint. By combining strict topological bipartition, frozen historical GDFs, orthogonal column masking, and bias freezing, CLEAN preserves historical diagnostic outputs pointwise without replaying old responses, while its expandable full-rank and low-rank branches retain strong plasticity for newly introduced concepts. Experiments on Math1, Junyi, and ASSISTments, representing dense, moderately sparse, and highly sparse response settings, respectively, show that CLEAN achieves zero representation drift ($\mathrm{RD}=0$) and maintains anchor-level performance on old items. CLEAN also achieves performance competitive with or superior to that of existing continual learning baselines. Under concept-space expansion, these results establish architectural isolation as an effective solution for eliminating catastrophic forgetting while maintaining psychometric consistency. Future work will extend CLEAN to multi-stage continual streams and investigate long-horizon consistency, cumulative capacity growth, and adaptation under repeated item and concept expansion.

\bibliographystyle{ACM-Reference-Format}
\bibliography{essay_references_no_doi_url}

\appendix

\section{Appendix  }

\subsection{User-Split Results}
\label{app:user-split}

\begin{table*}[t]
  \caption{User-Split results for score prediction.}
  \label{tab:user-split-score}
  \centering
  \small
  \setlength{\tabcolsep}{4pt}
  \renewcommand{\arraystretch}{1.15}

  \begin{tabular}{@{}lccccccc@{}}
    \toprule
    \textbf{Model}
    & \(\mathrm{ACC}_{\mathrm{old}}\uparrow\)
    & \(\mathrm{ACC}_{\mathrm{new}}\uparrow\)
    & \(\mathrm{F1}_{\mathrm{old}}\uparrow\)
    & \(\mathrm{F1}_{\mathrm{new}}\uparrow\)
    & \(\mathrm{RMSE}_{\mathrm{old}}\downarrow\)
    & \(\mathrm{RMSE}_{\mathrm{new}}\downarrow\)
    & \(\mathrm{RD}\downarrow\) \\
    \midrule

    G-NCDM (Anchor)
    & $\best{70.24\pm0.54}$
    & --
    & $\second{79.65\pm0.69}$
    & --
    & $\best{45.22\pm0.29}$
    & --
    & -- \\

    \midrule

    EWC
    & $68.73\pm0.77$
    & $65.48\pm0.58$
    & $76.90\pm0.61$
    & $73.06\pm0.62$
    & $49.69\pm0.54$
    & $51.52\pm0.32$
    & $8.47\pm0.12$ \\

    DER++
    & $67.33\pm0.28$
    & $65.29\pm0.57$
    & $75.90\pm0.18$
    & $72.97\pm0.58$
    & $50.09\pm0.26$
    & $51.08\pm0.33$
    & $11.60\pm0.06$ \\

    C-LoRA
    & $68.81\pm0.66$
    & $67.22\pm0.70$
    & $77.80\pm0.64$
    & $75.60\pm0.57$
    & $46.24\pm0.45$
    & $47.41\pm0.41$
    & $14.38\pm0.24$ \\

    X-DER
    & $\second{69.95\pm0.57}$
    & $66.87\pm0.54$
    & $\best{79.70\pm0.68}$
    & $77.15\pm0.55$
    & $\second{45.33\pm0.31}$
    & $46.39\pm0.15$
    & $\second{3.34\pm0.62}$ \\

    Full-Replay Oracle (reference)
    & $69.52\pm0.77$
    & $67.82\pm0.61$
    & $78.93\pm0.89$
    & $77.63\pm0.55$
    & $45.64\pm0.31$
    & $45.99\pm0.23$
    & $3.44\pm1.04$ \\

    C-LoRA-GNCDM
    & $62.90\pm1.97$
    & $64.74\pm2.26$
    & $71.06\pm2.49$
    & $71.81\pm3.41$
    & $50.10\pm1.56$
    & $47.40\pm0.74$
    & $3.77\pm0.72$ \\

    ICD
    & $51.34\pm0.00$
    & $63.39\pm0.00$
    & $52.62\pm0.00$
    & $71.31\pm0.00$
    & $50.38\pm0.00$
    & $48.89\pm0.00$
    & $412.16\pm0.00$ \\

    \midrule

    CLEAN-Full
    & $\best{70.24\pm0.54}$
    & $\second{68.30\pm0.74}$
    & $\second{79.65\pm0.69}$
    & $\second{77.65\pm0.91}$
    & $\best{45.22\pm0.29}$
    & $\second{45.42\pm0.20}$
    & $\best{0.00\pm0.00}$ \\

    CLEAN-LoRA
    & $\best{70.24\pm0.54}$
    & $\best{69.92\pm0.40}$
    & $\second{79.65\pm0.69}$
    & $\best{78.58\pm0.23}$
    & $\best{45.22\pm0.29}$
    & $\best{44.76\pm0.22}$
    & $\best{0.00\pm0.00}$ \\

    \bottomrule
  \end{tabular}

  \vspace{3pt}

  \begin{minipage}{0.98\textwidth}
    \footnotesize
    \raggedright
    \textit{Note.}
    Higher values are better for ACC and F1, whereas lower values are
    better for RMSE and RD. Bold and underlined values indicate the best
    and second-best means in each column, respectively. Tied results
    receive the same ranking.
  \end{minipage}
\end{table*}

Table~\ref{tab:user-split-score} reports the response-prediction results
on ASSISTments 2009--2010 under the User-Split setting, in which the
training, validation, and test sets contain disjoint groups of learners.
This protocol evaluates each model's ability to diagnose previously unseen
learners from a limited set of support responses. The table also includes
the original C-LoRA configuration as a reference for comparison with the
G-NCDM-based variant used in our experiments. The results show that
replacing the original backbone with G-NCDM does not by itself improve
predictive performance. C-LoRA-GNCDM generally performs worse than the
original C-LoRA on both old and new items, although their
$\mathrm{RMSE}_{\mathrm{new}}$ values are nearly identical.

All methods use the same support--query partition and are evaluated on the
same held-out query interactions. G-NCDM-based methods infer learner traits
directly from the support responses, whereas the transductive baselines use
these responses to optimize learner-specific embeddings, and ICD
incorporates the support interactions into its neighborhood graph.
Therefore, although the methods process support information differently,
their predictive performance is measured on the same query set.

CLEAN-LoRA achieves the best predictive performance on new items, with
$\mathrm{ACC}_{\mathrm{new}}=69.92$,
$\mathrm{F1}_{\mathrm{new}}=78.58$, and
$\mathrm{RMSE}_{\mathrm{new}}=44.76$, while maintaining zero
representation drift on historical concepts
($\mathrm{RD}=0.00$). It also exactly preserves the anchor-level
performance on old items, achieving
$\mathrm{ACC}_{\mathrm{old}}=70.24$,
$\mathrm{F1}_{\mathrm{old}}=79.65$, and
$\mathrm{RMSE}_{\mathrm{old}}=45.22$.
These results indicate that freezing the historical GDF preserves the
existing diagnostic space without preventing adaptation to newly
introduced items.

Among the non-CLEAN baselines, Full-Replay Oracle achieves an
$\mathrm{ACC}_{\mathrm{new}}$ of 67.82, which is 2.10 percentage points
lower than that of CLEAN-LoRA, while producing nonzero representation
drift ($\mathrm{RD}=3.44$). EWC, DER++, C-LoRA, X-DER,
C-LoRA-GNCDM, and ICD also exhibit nonzero representation drift and
generally weaker predictive performance on new items. In particular, the
larger drift observed for DER++, C-LoRA, and ICD suggests that replay,
parameter-efficient updating, and graph-based adaptation do not by
themselves guarantee the stability of the historical diagnostic space.

Overall, these results provide additional evidence that CLEAN-LoRA achieves
a favorable stability--plasticity trade-off for previously unseen learners
under limited-support conditions. It preserves historical representations
and old-item predictive performance while maintaining strong adaptability
to newly introduced items.

\subsection{Detailed Metric Results on Each Dataset}
\label{app:detailed-metrics}
Tables~\ref{tab:junyi_detailed}--~\ref{tab:math1_detailed} report the detailed ACC, F1, RMSE, and RD results on
Junyi-Random, ASSISTments a0910, and Math1, respectively.
Across all three datasets, CLEAN-Full and CLEAN-LoRA exactly match the
G-NCDM anchor in $\mathrm{ACC}_{\mathrm{old}}$,
$\mathrm{F1}_{\mathrm{old}}$, and $\mathrm{RMSE}_{\mathrm{old}}$,
while consistently achieving $\mathrm{RD}=0$.
This confirms that CLEAN preserves both historical trait representations
and old-item prediction performance after incremental updates.
On Junyi-Random, CLEAN-Full achieves an
$\mathrm{ACC}_{\mathrm{new}}$ of 74.53, an
$\mathrm{F1}_{\mathrm{new}}$ of 81.10, and an
$\mathrm{RMSE}_{\mathrm{new}}$ of 41.83, remaining close to the
Full-Replay Oracle without replaying historical data.
A similar pattern is observed on ASSISTments a0910, where CLEAN-Full
obtains 72.21, 79.61, and 43.45 for the three new-item metrics,
respectively.
On Math1, replay-based methods achieve stronger new-item performance,
whereas CLEAN-Full retains the anchor-level old-item metrics with zero
representation drift.
Overall, the detailed results demonstrate that CLEAN provides strict
historical stability across datasets while maintaining competitive
adaptability to newly introduced items and concepts.

\begin{table*}[t]
  \caption{Detailed metric results on the Junyi-Random dataset.}
  \label{tab:junyi_detailed}
  \centering
  \small
  \setlength{\tabcolsep}{4pt}
  \renewcommand{\arraystretch}{1.15}

  \begin{tabular}{@{}lccccccc@{}}
    \toprule
    \textbf{Model}
    & \(\mathrm{ACC}_{\mathrm{old}}\uparrow\)
    & \(\mathrm{ACC}_{\mathrm{new}}\uparrow\)
    & \(\mathrm{F1}_{\mathrm{old}}\uparrow\)
    & \(\mathrm{F1}_{\mathrm{new}}\uparrow\)
    & \(\mathrm{RMSE}_{\mathrm{old}}\downarrow\)
    & \(\mathrm{RMSE}_{\mathrm{new}}\downarrow\)
    & \(\mathrm{RD}\downarrow\) \\
    \midrule

    G-NCDM (Anchor)
    & \underline{78.40$\pm$0.15}
    & --
    & \underline{85.53$\pm$0.14}
    & --
    & \underline{38.72$\pm$0.05}
    & --
    & -- \\

    \midrule

    EWC
    & 71.54$\pm$2.03
    & 71.58$\pm$0.40
    & 79.18$\pm$2.05
    & 78.27$\pm$0.52
    & 44.83$\pm$1.34
    & 43.98$\pm$0.21
    & 8.92$\pm$0.20 \\

    DER++
    & 76.53$\pm$0.28
    & 74.27$\pm$0.26
    & 83.66$\pm$0.24
    & 80.44$\pm$0.25
    & 40.00$\pm$0.25
    & \underline{41.61$\pm$0.13}
    & \underline{3.09$\pm$1.14} \\

    C-LoRA
    & 71.26$\pm$2.01
    & 73.12$\pm$0.36
    & 79.41$\pm$1.74
    & 79.32$\pm$0.34
    & 44.10$\pm$1.31
    & 42.93$\pm$0.21
    & 19.69$\pm$0.27 \\

    X-DER
    & 77.59$\pm$0.42
    & 73.19$\pm$0.69
    & 85.28$\pm$0.21
    & 79.73$\pm$0.50
    & 39.60$\pm$0.32
    & 42.54$\pm$0.24
    & 6.36$\pm$1.06 \\

    Full-Replay Oracle (reference)
    & \textbf{78.55$\pm$0.25}
    & \textbf{75.47$\pm$0.22}
    & \textbf{85.74$\pm$0.17}
    & \textbf{81.49$\pm$0.26}
    & \textbf{38.71$\pm$0.12}
    & \textbf{40.86$\pm$0.09}
    & 4.88$\pm$1.29 \\

    ICD
    & 70.01$\pm$1.47
    & 66.84$\pm$4.50
    & 70.74$\pm$8.29
    & 64.20$\pm$13.46
    & 45.27$\pm$1.64
    & 47.61$\pm$0.44
    & 119.02$\pm$24.52 \\

    \midrule

    CLEAN-Full
    & \underline{78.40$\pm$0.15}
    & \underline{74.53$\pm$0.30}
    & \underline{85.53$\pm$0.14}
    & \underline{81.10$\pm$0.13}
    & \underline{38.72$\pm$0.05}
    & 41.83$\pm$0.11
    & \textbf{0.00$\pm$0.00} \\

    CLEAN-LoRA
    & \underline{78.40$\pm$0.15}
    & 73.57$\pm$0.64
    & \underline{85.53$\pm$0.14}
    & 80.65$\pm$0.39
    & \underline{38.72$\pm$0.05}
    & 42.43$\pm$0.44
    & \textbf{0.00$\pm$0.00} \\

    \bottomrule
  \end{tabular}

\end{table*}

\begin{table*}[t]
  \caption{Detailed metric results on the ASSISTments-Random dataset.}
  \label{tab:a0910_detailed}
  \centering
  \small
  \setlength{\tabcolsep}{4pt}
  \renewcommand{\arraystretch}{1.15}

  \begin{tabular}{@{}lccccccc@{}}
    \toprule
    \textbf{Model}
    & \(\mathrm{ACC}_{\mathrm{old}}\uparrow\)
    & \(\mathrm{ACC}_{\mathrm{new}}\uparrow\)
    & \(\mathrm{F1}_{\mathrm{old}}\uparrow\)
    & \(\mathrm{F1}_{\mathrm{new}}\uparrow\)
    & \(\mathrm{RMSE}_{\mathrm{old}}\downarrow\)
    & \(\mathrm{RMSE}_{\mathrm{new}}\downarrow\)
    & \(\mathrm{RD}\downarrow\) \\
    \midrule

    G-NCDM (Anchor)
    & \underline{73.51$\pm$0.15}
    & --
    & \underline{81.35$\pm$0.26}
    & --
    & \textbf{42.68$\pm$0.12}
    & --
    & -- \\

    \midrule

    EWC
    & 66.87$\pm$0.67
    & 64.09$\pm$0.61
    & 74.79$\pm$0.91
    & 71.84$\pm$0.72
    & 52.50$\pm$0.38
    & 53.88$\pm$0.41
    & 9.01$\pm$0.09 \\

    DER++
    & 69.50$\pm$0.44
    & 67.66$\pm$0.51
    & 77.74$\pm$0.53
    & 76.34$\pm$0.40
    & 45.08$\pm$0.42
    & 46.25$\pm$0.26
    & \underline{3.64$\pm$0.94} \\

    C-LoRA
    & 67.64$\pm$0.28
    & 64.28$\pm$0.46
    & 75.79$\pm$0.35
    & 72.01$\pm$0.48
    & 52.13$\pm$0.18
    & 53.94$\pm$0.35
    & 13.11$\pm$0.09 \\

    X-DER
    & 72.86$\pm$0.24
    & 69.83$\pm$0.44
    & 81.18$\pm$0.28
    & 78.26$\pm$0.24
    & 43.02$\pm$0.21
    & 44.79$\pm$0.26
    & 7.32$\pm$1.57 \\

    Full-Replay Oracle (reference)
    & \textbf{73.66$\pm$0.12}
    & \textbf{72.99$\pm$0.33}
    & \textbf{81.58$\pm$0.21}
    & \textbf{80.20$\pm$0.45}
    & \underline{42.77$\pm$0.10}
    & \textbf{43.08$\pm$0.12}
    & 6.92$\pm$0.72 \\

    ICD
    & 54.40$\pm$0.00
    & 65.00$\pm$0.00
    & 59.20$\pm$0.00
    & 75.10$\pm$0.00
    & 49.16$\pm$0.00
    & 47.28$\pm$0.00
    & 422.42$\pm$0.00 \\

    \midrule

    CLEAN-Full
    & \underline{73.51$\pm$0.15}
    & \underline{72.21$\pm$0.19}
    & \underline{81.35$\pm$0.26}
    & \underline{79.61$\pm$0.20}
    & \textbf{42.68$\pm$0.12}
    & \underline{43.45$\pm$0.11}
    & \textbf{0.00$\pm$0.00} \\

    CLEAN-LoRA
    & \underline{73.51$\pm$0.15}
    & 72.12$\pm$0.24
    & \underline{81.35$\pm$0.26}
    & 79.55$\pm$0.40
    & \textbf{42.68$\pm$0.12}
    & 43.80$\pm$0.23
    & \textbf{0.00$\pm$0.00} \\

  \bottomrule
  \end{tabular}
\end{table*}

\begin{table*}[t]
  \caption{Detailed metric results on the Math1-Random dataset.}
  \label{tab:math1_detailed}
  \centering
  \small
  \setlength{\tabcolsep}{4pt}
  \renewcommand{\arraystretch}{1.15}

  \begin{tabular}{@{}lccccccc@{}}
    \toprule
    \textbf{Model}
    & \(\mathrm{ACC}_{\mathrm{old}}\uparrow\)
    & \(\mathrm{ACC}_{\mathrm{new}}\uparrow\)
    & \(\mathrm{F1}_{\mathrm{old}}\uparrow\)
    & \(\mathrm{F1}_{\mathrm{new}}\uparrow\)
    & \(\mathrm{RMSE}_{\mathrm{old}}\downarrow\)
    & \(\mathrm{RMSE}_{\mathrm{new}}\downarrow\)
    & \(\mathrm{RD}\downarrow\) \\
    \midrule

    G-NCDM (Anchor)
    & 72.14$\pm$0.51
    & --
    & 71.72$\pm$1.65
    & --
    & 42.43$\pm$0.42
    & --
    & -- \\

    \midrule

    EWC
    & 68.59$\pm$0.51
    & 73.27$\pm$0.86
    & 66.01$\pm$1.63
    & 61.64$\pm$1.55
    & 50.30$\pm$0.45
    & 43.11$\pm$0.68
    & 9.70$\pm$0.26 \\

    DER++
    & 71.72$\pm$0.52
    & \textbf{75.52$\pm$0.22}
    & \underline{72.87$\pm$1.36}
    & \textbf{67.02$\pm$1.14}
    & 42.56$\pm$0.36
    & \textbf{39.00$\pm$0.10}
    & \underline{1.32$\pm$1.11} \\

    C-LoRA
    & 68.88$\pm$0.51
    & 71.98$\pm$0.68
    & 66.45$\pm$1.61
    & 59.86$\pm$1.08
    & 50.28$\pm$0.46
    & 46.05$\pm$0.58
    & 10.82$\pm$0.27 \\

    X-DER
    & \textbf{72.54$\pm$0.21}
    & \underline{75.12$\pm$0.35}
    & \textbf{73.45$\pm$1.47}
    & 66.10$\pm$1.02
    & \textbf{42.15$\pm$0.34}
    & \underline{39.07$\pm$0.19}
    & 7.24$\pm$2.00 \\

    Full-Replay Oracle (reference)
    & \underline{72.20$\pm$0.28}
    & \underline{75.12$\pm$0.28}
    & 71.94$\pm$0.85
    & \underline{66.63$\pm$0.98}
    & \underline{42.18$\pm$0.31}
    & 39.66$\pm$0.28
    & 6.47$\pm$1.60 \\

    ICD
    & 68.62$\pm$0.00
    & 71.10$\pm$0.00
    & 62.88$\pm$0.00
    & 51.43$\pm$0.00
    & 46.82$\pm$0.00
    & 47.19$\pm$0.00
    & 95.75$\pm$0.00 \\

    \midrule

    CLEAN-Full
    & 72.14$\pm$0.51
    & 72.25$\pm$4.69
    & 71.72$\pm$1.65
    & 57.53$\pm$2.75
    & 42.43$\pm$0.42
    & 45.64$\pm$2.39
    & \textbf{0.00$\pm$0.00} \\

    CLEAN-LoRA
    & 72.14$\pm$0.51
    & 68.50$\pm$3.00
    & 71.72$\pm$1.65
    & 54.73$\pm$2.50
    & 42.43$\pm$0.42
    & 47.73$\pm$1.59
    & \textbf{0.00$\pm$0.00} \\

    \bottomrule
  \end{tabular}

  \vspace{3pt}

  \begin{minipage}{0.98\textwidth}
    \footnotesize
    \raggedright
    \textit{Note.}
    All values are reported as percentages (\%) in the form of mean
    $\pm$ sample standard deviation over 10 random seeds.
    Higher values are better for ACC and F1, whereas lower values are
    better for RMSE and RD.
    Bold and underlined values indicate the best and second-best means
    in each column, respectively.
    The Anchor model participates only in the ranking of historical-item
    metrics.
    Full-Replay Oracle may replay all historical training data during
    each incremental stage and employs full-parameter expansion, serving
    as a theoretical performance upper bound.
  \end{minipage}
\end{table*}

\subsection{Sensitivity Analysis of the Residual Mixing Coefficient
\texorpdfstring{$\alpha$}{alpha}}
\label{app:alpha-sensitivity}
\begin{figure}[H]
  \centering
  \includegraphics[width=\columnwidth]{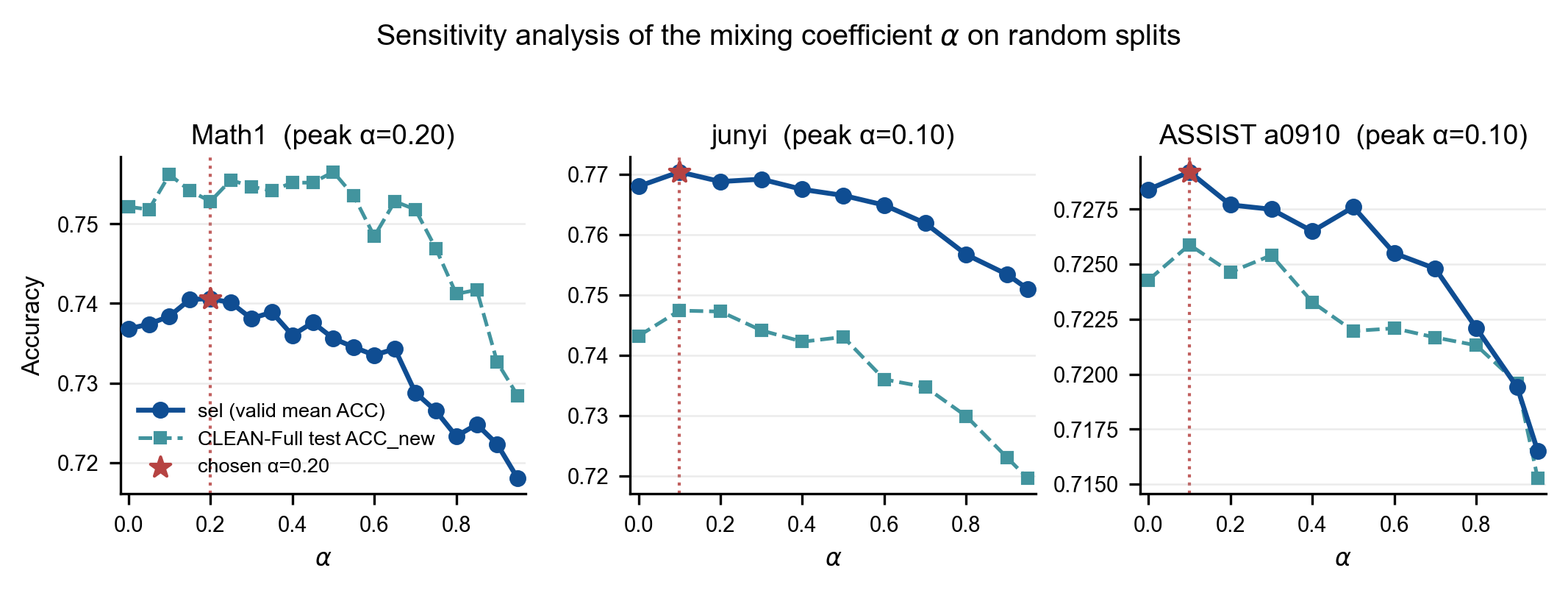}
  \caption{Sensitivity analysis of the coefficient $\alpha$ under the Random-Split setting.}
  \Description{Sensitivity analysis of the coefficient alpha under the Random-Split setting.}
  \label{fig:alpha_sensitivity_random}
\end{figure}

The coefficient $\alpha$ balances the implicitly learned representation
and the explicitly extracted $Q$-matrix representation. A smaller
$\alpha$ places greater weight on $\theta^{\mathrm{imp}}$, whereas a
larger $\alpha$ emphasizes $\theta^{\mathrm{exp}}$. We evaluate different values of  under the Random-Split setting. As shown in the left panel of Figure ~\ref{fig:alpha_sensitivity_random}, the mean validation accuracy on Math1 first increases and then decreases, reaching its maximum at $\alpha$=0.2. For Junyi and ASSIST09-10, the best performance is achieved at $\alpha$=0.1, after which accuracy declines as  increases. A similar downward trend is observed in accuracy on incremental data, particularly at large values of . These results indicate that excessive reliance on foundation-stage information limits the model’s ability to adapt to incremental data. Although the optimal value varies across datasets, relatively small values of  consistently provide a better trade-off between retaining previously learned information and incorporating new information. Based on the validation results, we set $\alpha$=0.2 for Math1 and $\alpha$=0.1 for Junyi and ASSIST09-10.

\end{document}